# Open Source 3-D Filament Diameter Sensor for Recycling, Winding and Additive Manufacturing Machines

Aliaksei L. Petsiuk[1] and Joshua M. Pearce[1,2,3]

[1]Department of Electrical & Computer Engineering, Michigan Technological University, Houghton, MI 49931, USA
[2]Department of Material Science & Engineering, Michigan Technological University, Houghton, MI 49931, USA
[3]Department of Electronics and Nanoengineering, School of Electrical Engineering, Aalto University, Espoo, FI-00076, Finland

apetsiuk@mtu.edu, pearce@mtu.edu

**Graphical Abstract**

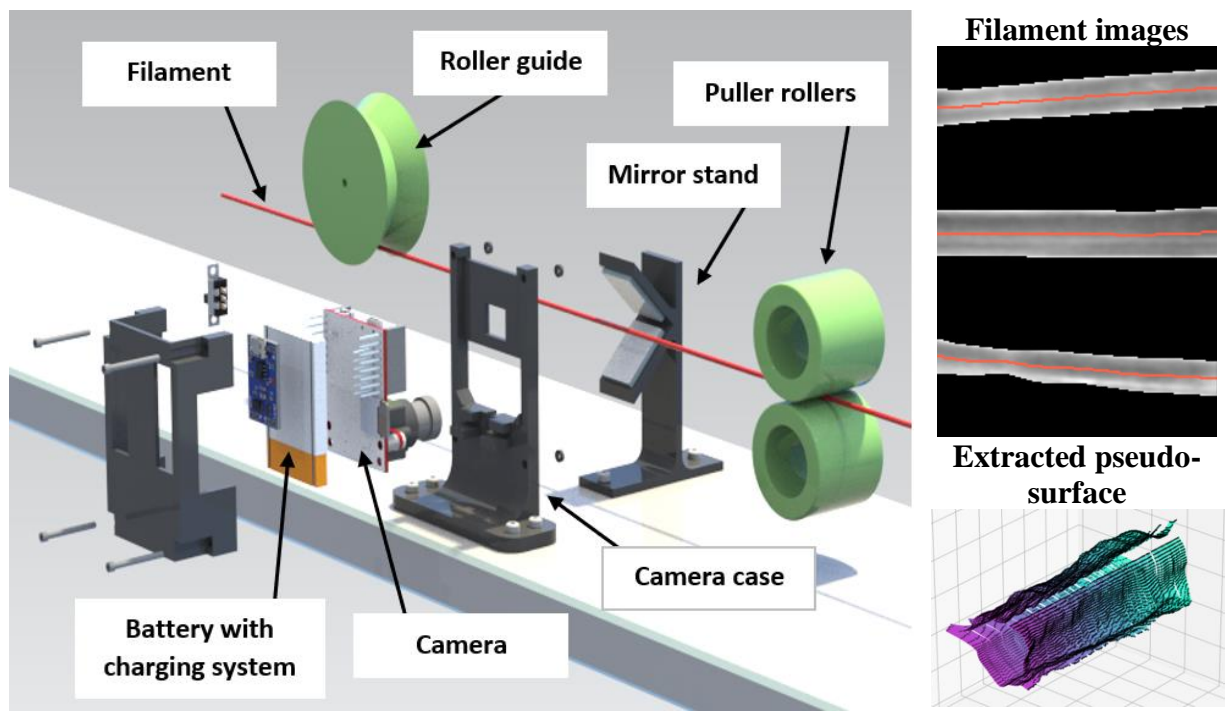

**Filament images**

**Extracted pseudo-surface**

**Highlights**

- Developed a portable optical computer vision-based diameter measurement system.
- The module can be integrated into a control system of a plastic waste recycling machine.
- Multi-axis measurements with ovality detection and texture analysis assure quality.
- Measurement logging allows tracking the quality of the filament along its length.

**Abstract**

To overcome the challenge of upcycling plastic waste into 3-D printing filament in the distributed recycling and additive manufacturing systems, this study designs, builds, tests and validates an open source 3-D filament diameter sensor for recycling and winding machines. The modular system for multi-axis optical control of the diameter of the recycled 3-D-printer filament makes it possible to analyze the surface structure of the processed filament, save the history of measurements along the entire length of the spool, as well as mark defective areas. The sensor is developed as an independent module and integrated into a recyclebot. The diameter sensor was tested on different kinds of polymers (ABS, PLA) different sources of plastic (recycled 3-D prints and virgin plastic waste) and different colors including clear plastic. The results of the diameter measurements using the camera were compared with the manual measurements, and the measurements obtained with a one-dimensional digital light caliper. The results found that the developed open source filament sensing method allows users to obtain significantly more information in comparison with basic one-dimensional light sensors and using the received data not only for more accurate diameter measurements, but also for a detailed analysis of the recycled filament surface. The developed method ensures greater availability of plastics recycling technologies for the manufacturing community and stimulates the growth of composite materials creation. The presented system can greatly enhance the user possibilities and serve as a starting point for a complete recycling control system that will regulate motor parameters to achieve the desired filament diameter with acceptable deviations and even control the extrusion rate on a printer to recover from filament irregularities.

## 1. Introduction

Since the 1950s, only 9% of the plastic produced was recycled (Geyer et al., 2017) and this challenge was recently exacerbated (Joyce, 2019) when the world's leading plastic recycler, China (McNaugton, 2019), imposed an import ban on waste plastic (Brooks et al., 2018) and stalled global recycling (Katz, 2019). The market for large-scale centralized plastic recycling is thus often uneconomic, and many areas ceased plastic recycling entirely (Corkery, 2019). With an exponential growth, the world produces about 300 million tons of plastic waste annually (Geyer et al., 2017), from 4 to 12 million of which contaminate the natural environment often ending up in the ocean (Jambeck et al., 2015). Complicating the plastic recycling challenge is the explosive popularity of 3-D printing, which played a role the plastics market growth over the past decade (Report Buyer, 2019). According to the economic forecasts, the global 3-D printing market is to increase from $2.1 billion in 2018 to $7.7 billion in 2024, being driven by end-user awareness of 3-D printing technology and its increasing use across oil and gas industries (Globe Newswire, 2019). The most common plastics used in the most popular type of 3-D printing, fused filament fabrication technology (FFF), are acrylonitrile butadiene styrene (ABS), polylactide (PLA), and polyethylene terephthalate glycol (PETG) due to their availability and printability (Report Buyer, 2019). ABS is a thermoplastic based on elastomers that make it flexible and shock resistant. It holds the largest global 3-D printing filament market share. Recent analysis reveals a shift towards biodegradable PLA based on the use of renewable materials like corn starch as the FFF material of choice in the future (Carlota, 2019). Rapid growth of the 3-D printing industry and particularly distributed manufacturing on low-cost FFF systems based on the open source self-replicating rapid prototyper (RepRap) model (Sells et al., 2009), where the 3-D printer fabricates its own parts (Jones et al., 2011), increases the potential for exponential growth of such machines (Bowyer, 2014). This provides a strong reason for concern of recycling the resultant plastic waste.

Although distributed manufacturing with FFF tends to have a lower environmental impact than central recycling (Kreiger and Pearce, 2013a) largely due to materials use and embodied transportation (Kreiger and Pearce, 2013b), the availability and breadth of 3-D printing technology stimulate experimental design and innovations, while significantly ramping up the number of defective parts and waste products (Faludi et al., 2017). Similarly, although 3-D printing creates little to no waste when compared to subtractive manufacturing, it is hard to assess the real statistics of failures since, in addition to printing errors that are relatively uncommon, there are also design mistakes and the potential for increased consumption. The latter would be due to the reduction in cost to manufacture plastic products for development (Gwamuri et al., 2014) or common household consumer goods (Wittbrodt et al., 2013) even made by consumers (Petersen and Pearce, 2017) including children for their own toys (Petersen et al. 2017). The amount of plastic waste generated by 3-D printers in user workplaces can be substantial as for example the University of California at Berkeley, produced 212 kilograms by their own set of a hundred of printers in 2017 (Barrett, 2020). As per a survey made by Filamentive, 8,000 tons of printed material were dumped in landfills worldwide in 2019 (Barrett, 2020).

Thus, today there are many independent groups of developers involved in the creation of devices for the recycling of plastic waste (Obudho, 2018). The ASTM (American Society for Testing and Materials) International Resin Identification Coding System classifies both ABS and PLA as the types of plastic waste that are not typically processed by municipal programs (Jones, 2019). The amount of recycled material is increasing by 0.7% per year since 1987 (Ritchie and Roser, 2018). Thus, in 2015, with an annual output of 380 million tons, the recycling of plastic waste was only 20%, and by 2050, recycling up to 44% is predicted (Ritchie and Roser, 2018). Being a critically important action to reduce environmental pollution, recycling represents a dynamic area in the modern plastic industry (Hopewell et al., 2009).

A rapidly growing approach to increase recycling rates and particularly those from the additive manufacturing industry to reach a circular economy for plastic (Zhong and Pearce, 2018) is the distributed recycling for additive manufacturing (DRAM) approach (Sanchez et al. 2020), which is profitable for participants (Pavlo et al., 2018). Unlike the conventional recycling model, consumers have an economic incentive (Zhong and Pearce, 2018) based on savings (Sanchez et al. 2020) to recycle with DRAM. This is because they can use their waste as 3-D printing feedstocks, which are a relatively high value of plastic (e.g. being sold for $20/kg or more). DRAM, thus has the potential to radically impact global value chains (Laplume et al., 2016).

The majority of DRAM research was centered on open source waste plastic extruders known as recyclebots, which upcycle post-consumer plastic waste into 3-D printing filament made from machined (Baechler et al., 2013) or 3-D printed parts (Woern et al., 2018a). In addition to reducing 3-D printing material costs below $0.10/kg, it decreased embodied energy of 3-D printing filament by 90% and thus reduced environmental impact (Kreiger et al., 2013) based on life cycle analysis (Kreiger et al., 2014) particularly when solar powered (Zhong et al., 2017). Using a recyclebot process, the recycling of plastic waste consists of three stages: 1) shredding used printed parts or

post-consumer waste plastic into small granules, 2) melting and extruding them into filaments, and 3) 3-D printing the filament into new products (Dertinger et al., 2020). There are several commercial shredders and extruders available on the market (Obudho, 2018) such as SHR3D IT (3devo, 2020), Filabot (Filabot, 2020), Filastruder (Filastruder, 2020), ProtoCycler (ReDeTec, 2020), as well as open-source versions like the open source waste plastic granulator (Ravindran et al., 2019), Plastic Bank (Plastic Bank Extruder v1.0, 2020), Precious Plastic (Preciousplastic, 2020), the RepRapable Recyclebot (Woern et al, 2018a) and others. Regardless of the equipment used for filament extrusion, the control of a filament diameter consistency is a key stage in the recycling process (Cardona et al., 2016). Since 3-D printing software generates material feed rate through G-Code based on filament diameter, even small deviations from the nominal value can lead to over- or under extrusion, nozzle blockage, and irregular gaps in printed lines (Le Bouthillier, 2016). Thus, the diameter tolerance is the main quality indicator of the filament being produced. Among industrial manufacturers, the standard is a deviation of ±0.05 mm or a maximum allowable error of 0.1mm, which can be considered a small value given the length of the entire spool (Cardona et al., 2016). In industry, laser multi-axis contactless meters are used to control the diameter of the winding filament. This allows measuring from different angles and tracking the ovality and diameter deviations (Huber et al., 2020). The independent open source community often cannot afford to use such expensive tools. Therefore, measurement of the output filament diameter is either not used at all leading to inferior filament, or simple one-dimensional optical sensors are used (Woern et al, 2018a), which cannot provide comprehensive data on the nature of the spooled filament.

To overcome this challenge of upcycling plastic waste into 3-D printing filament using a DRAM approach this study designs, builds, tests and validates an open source 3-D filament diameter sensor for recycling and winding machines. The open source 3-D filament diameter sensor is developed as an independent module and integrated into the RepRapable Recyclebot (Woern et al., 2018a), which can process 400 g/hour of plastic filament with an output diameter consistency of ±4.6%. The diameter sensor was tested on different kinds of polymers (ABS, PLA) different sources of plastic (recycled 3-D prints and virgin plastic waste) and different colors including clear plastic. The results of the diameter measurements using the camera were compared with the manual measurements, and the measurements obtained with a one-dimensional digital light caliper (Woern et al., 2018a). The results are discussed in the context of DRAM.

## 2. Methods
### 2.1 Mechanical Design
The developed design of the optical system (Figure 1) is based on OpenMV H7 camera (OpenMV, 2020), which includes a microcontroller board running on MicroPython (MicroPython, 2020) operating system. The 480 MHz ARM processor has USB, I2C, SPI interfaces, and three I/O pins for servo control. The availability of various interfaces allows the camera to be used as an independent motor control module. The power consumption of the entire system is 150 mA, allowing the camera to operate continuously for 6 hours using a compact 1000 mAh lithium-ion battery.

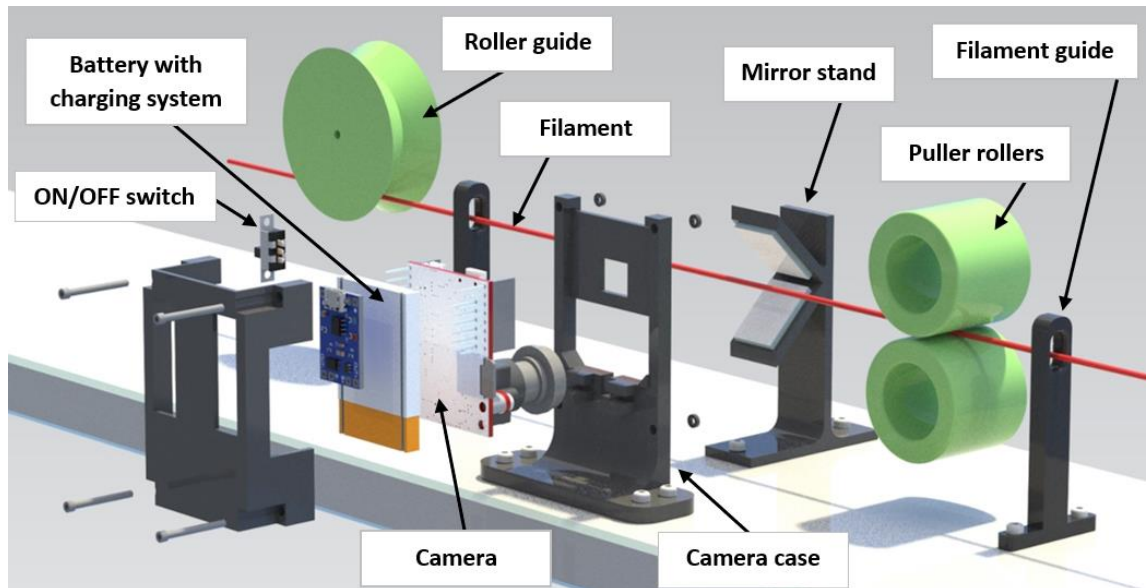

**Figure 1. Schematic view of the open source single camera-based optical diameter measuring system**

The bill of materials is presented in Table 1. All printed parts files are available in the OSF repository (Petsiuk and Pearce, 2020). It requires about 50 grams of plastic to print the necessary parts. The total cost of the module is almost entirely determined by the price of the camera and battery and is no more than US $80. It should be noted that the camera costs fall steeply with volume purchases (e.g. by 10% for 10 and 30% for 100 on openmv.io). Depending on the design and tasks of the system, the camera module can operate both from a portable battery and from a separate power supply or from a common power rail in case of a joint operation with the electrical system of the recyclebot.

**Table 1. Bill of materials**

| No. | Component | Quantity | Mass of polymer filament |
|-----|-----------|----------|--------------------------|
| 1 | OpenMV H7 Camera | 1 | — |
| 2 | 3.7V 1000mAh lithium polymer battery | 1 | — |
| 3 | 5V 1A TP4056 battery charger | 1 | — |
| 4 | Slide switch | 1 | — |
| 5 | Filament guide | 2 | 6 g |
| 6 | Mirror stand | 1 | 18 g |
| 7 | Front camera case | 1 | 15 g |
| 8 | Back camera case | 1 | 14 g |
| 9 | Mirror plate | 2 | — |
| 10 | M3 wood screw | 8 | — |
| 11 | M2 screw | 4 | — |
| 12 | M2 nut | 4 | — |

The source file repository (Petsiuk and Pearce, 2020) contains the open-source code for the camera, written in Python, taking into account the specifics of the MicroPython environment, as well as Python scripts for post-processing of captured frames.

The camera source code is written to the Secure Digital (SD) memory card and starts working immediately after the power is turned on. A cyclic subroutine captures a frame and saves it to the SD card at a specified time interval. After winding, images of the filament and its specular reflections along the entire length of the spool are stored on the memory card. By changing the recording time interval, it is possible to track the entire length of the spool, as well as its individual sections.

The resulting series of images is then processed by the Python notebooks, where the filament is segmented, its diameter is measured pixel by pixel, and the visible parts of the surface are combined into texture patches for subsequent analysis. Knowing the filament speed and the frame capture period, it is possible to build a history of measurements, where each section of the filament is characterized by its own values of diameter, ovality, and deviations of surface texture.

Figure 2 illustrates the electrical wiring diagram. The camera with the battery and charging system is an independent module and is not connected to the main control system of the recyclebot. OpenMV camera, however, has the necessary interfaces for connecting additional sensors and provides the ability to directly control electric motors.

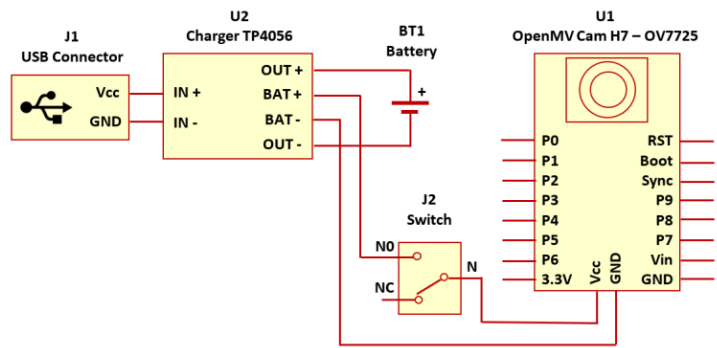

**Figure 2. Wiring diagram**

In the future, it is planned to expand the functionality of the optical module by adding adaptive winding control. As can be seen from Figure 3, with a uniform filament supply, the spool winding speed should decrease stepwise with an increase in the number of wound layers.

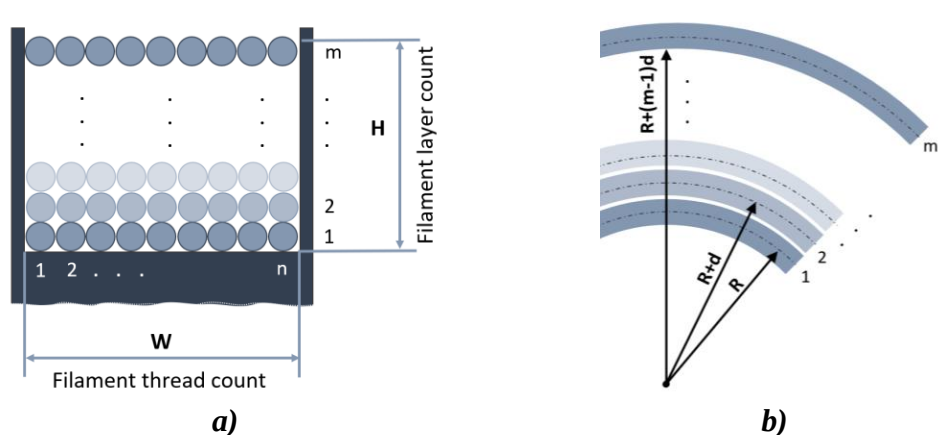

*a)*                                        *b)*

**Figure 3. Spool winding diagram:** a) schematic front view; b) schematic side view.

The length of the filament required to wind one layer is $L_m = 2\pi \cdot n \cdot (R+(m\text{-}1)\cdot d)$ millimeters, while the total length of the filament in the spool is $L_{Total}$ millimeters and can be expressed as follows:

$$L_{Total} = 2\pi \cdot n \cdot \left(R \cdot m + \sum_{i=2}^{m}(i-1)\cdot d\right) \qquad [\text{mm}] \quad (1)$$

Where $R$ is the spool radius in millimeters, $m$ is the number of filament layers, $n$ is the number of diameters that fit into the spool width, and $d$ is the filament diameter in millimeters.

As can be seen from the expressions above, the length of the wound filament and, therefore, the speed of its spooling, directly depend on the diameter $d$. The size of the diameter, in turn, depends on the feed rate of the solidifying plastic. The diameter of the filament, therefore, is a key parameter for the plastic recycling machine, and its inclusion in a feedback loop could automate the operation of the whole system.

## 2.2 Image Acquisition and Processing

The camera records 640x480 RGB images at 1 frame per second. Considering the extruding speed during the spooling process, it can capture the filament along its entire length without leaving unmeasured sections. The presence of mirrors above and below the filament, located at 45-degree angles, makes it possible to obtain images from three different sides (Figure 4). Thus, in addition to the triple measurement of the diameter, the cross-sectional ovality deviations are also calculated. It should be noted that the camera image is reversed upside-down, so the upper mirror in the image is the lower mirror in the setup.

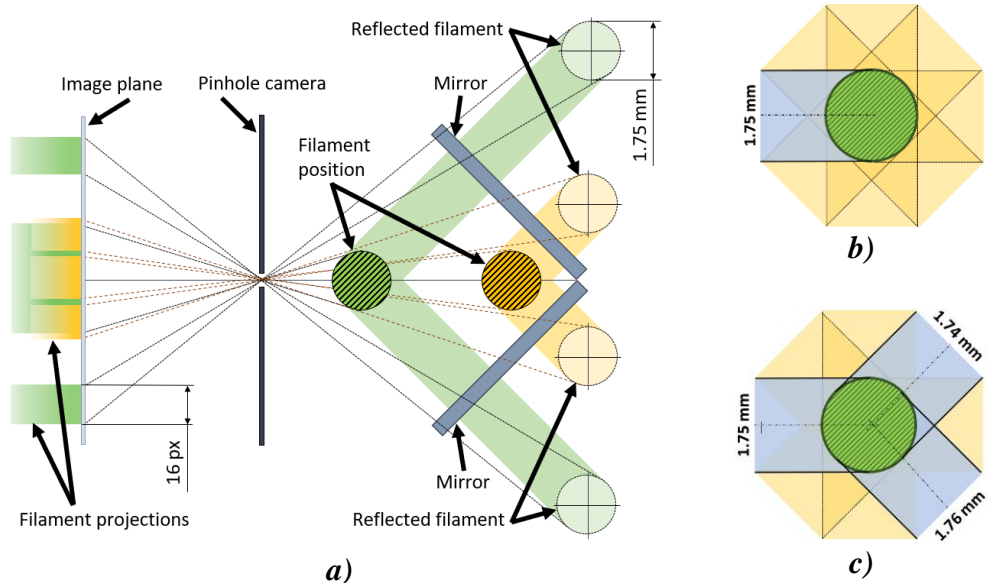

**Figure 4. Image acquisition scheme:** a) image acquisition based on the pinhole camera model; b) filament projection in the absence of mirrors; c) filament projection with two mirrors installed at 45-degree angles

According to the pinhole camera model (Figure 5), the size of the object projection $p$ depends on the distance to the camera $L$ and is equal to $p=F \cdot W/L$, where $p$ is the projection size in mm, $F$ is the constant focal length in mm, $W$ is the actual object size in the world coordinate system, $L$ – distance to the camera in mm, $O$ – camera location.

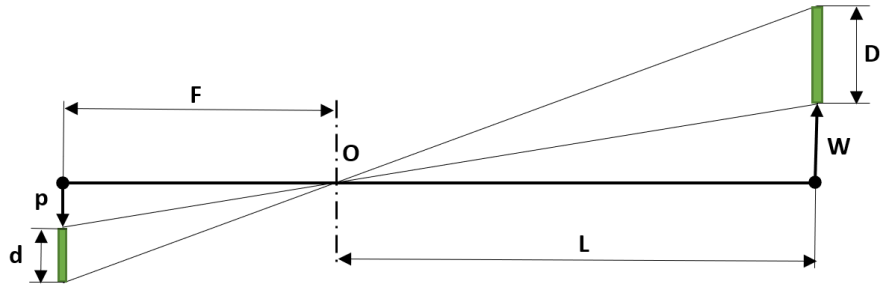

**Figure 5. Pinhole camera model**

The projection size is inversely proportional to the distance from the object to the camera and depends on lens parameters. Even without a separate camera calibration (no known camera focal length and lens distortion), however, it is possible to calibrate the entire optical system so that the measured diameter $D$ in millimeters is proportional to the projection size $d$ in pixels according to the scaling factor $s$: $D=d·s$, where $s=f(L)$ is a function of the distance to the camera. This calibration procedure is described in the next section.

Figure 6 illustrates the sequence of processing camera images using one frame as an example. Due to a significant amount of minor details in the image frame, as well as due to external reflections and highlights created by mirrors, regions of interest for the main image of the filament and for its two reflections were determined at the preliminary testing stage, which allows for the most consistent results.

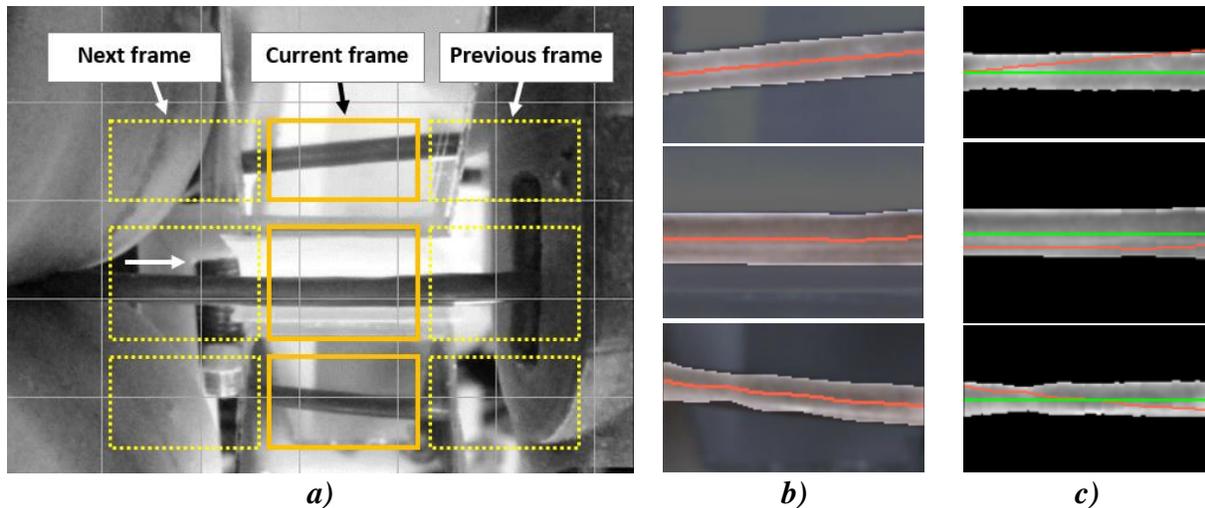

**Figure 6. Filament segmentation and diameter measurement procedure:** a) single camera frame; b) segmented filament; c) rectified regions reveal filament thickness and its surface structure based on light intensity. The red lines represent the detected filament centerlines, while the green ones represent their straightened counterparts, making it easier to concatenate adjacent texture sections together.

The segmentation is then performed based on the grayscale gradient for each single pixel slice of the region of interest. Since the boundaries of the filament image are characterized by a sharp change in the gray level value, the only variable segmenting parameter is the gradient threshold (Figure 7). In this way, the edges of the plastic filament are determined regardless of its color and material. It should be noted, however, that the color of the substrate on the mirror stand can be changed in advance to achieve greater contrast with the color and tone of the extruding filament.

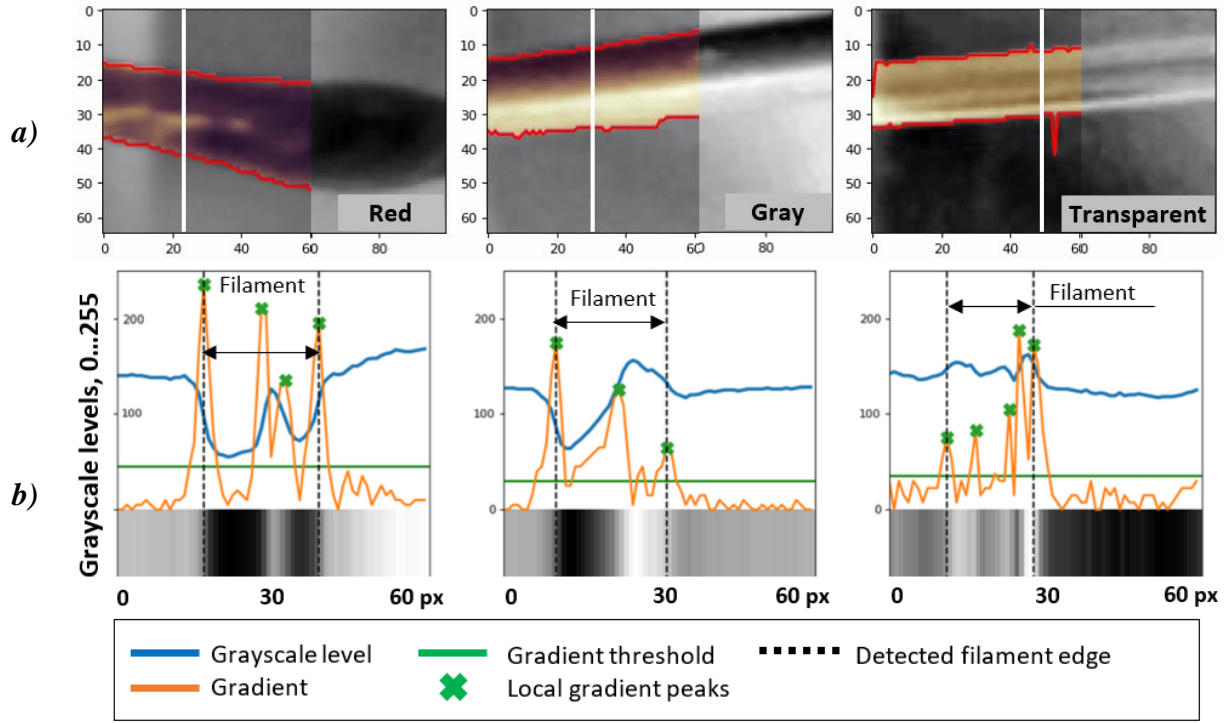

**Figure 7. Edge segmentation process:** a) grayscale input images with the selected single-pixel slices (white vertical bars); b) filament edges detected for the single-pixel image slices demonstrated for red, gray and transparent filament.

The collection of single-pixel patches (Figure 7) defines the masks of the entire filament section for all three projections. The segmented sections are rectified based on the corresponding midlines obtained by processing of the binary pixel masks found in the previous step.

All three regions of interest are $N$ pixels wide, which corresponds to $3 \cdot N$ diameter measurements per frame. Knowing the frame number and the extrusion rate, it is possible to keep the measurement history and label the section of produced filament in case of critical deviations.

Figure 8 shows a set of single-pixel vertical slices of a rectified filament section. The widths of the segmented areas represent the diameter of the filament, expressed in pixels, while the intensities of the reflected light (blue dots in the graphs) represent the surface character of the visible part of the filament. As a result, three projections of one filament section are obtained, which, in addition to the diameter, also provides information on the surface structure of the entire section.

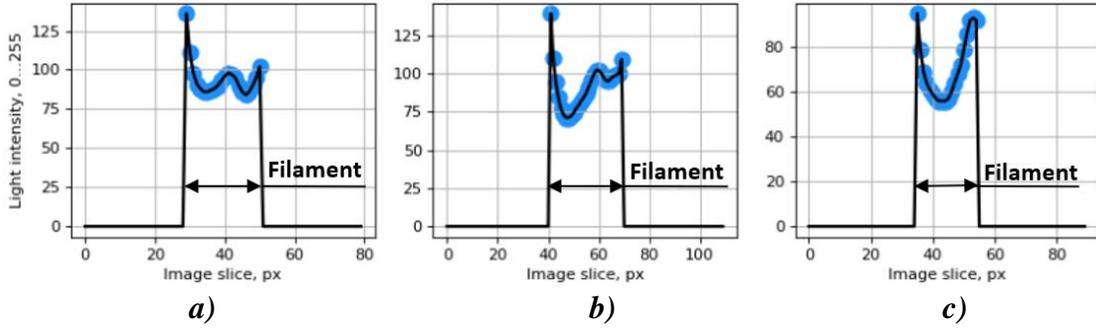

**Figure 8. Vertical slices of the segmented filament section:** a) upper mirror projection; b) main image; c) lower mirror projection.

For each single-pixel slice, the ovality is determined by:

$$O = \frac{D_{MAX} - D_{MIN}}{D_N} \cdot 100 \text{ %} \qquad [\%] \ (2)$$

Where $O$ is the percentage of ovality, $D_{MAX}$ and $D_{MIN}$ are the maximum and minimum detected diameters, respectively, and $D_N$ is the nominal diameter of the recycled filament.

The filament projections from three different angles (Figure 8) represent its full surface as a light intensity map (pseudo-surface). Figure 9, as an example, shows a single concatenated slice of the light intensity distribution on the filament surface in the direction perpendicular to its length.

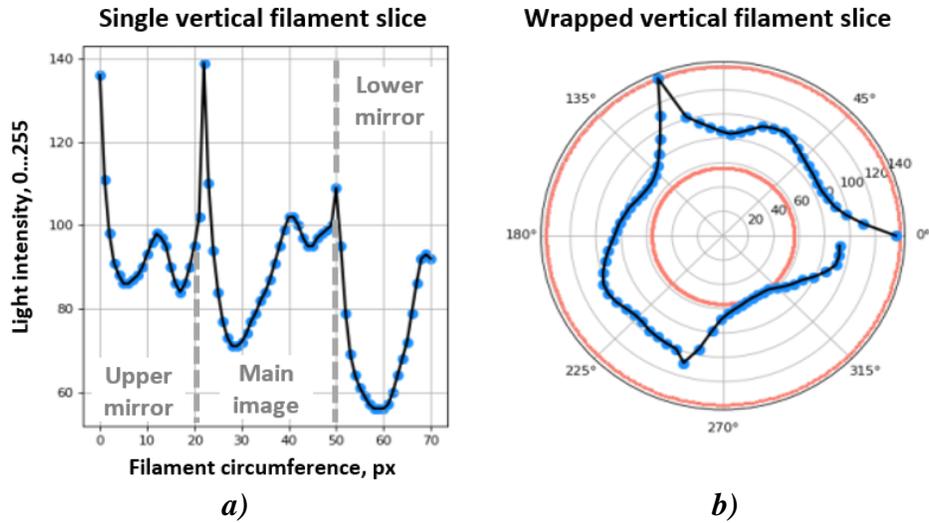

**Figure 9. Filament surface based on reflected light intensity:** a) a slice of a concatenated segmented filament image in a direction perpendicular to its length; b) cross-section of a wrapped pseudo-surface

Each pixel of the region of interest (Figure 6) forms a single vertical slice. Thus, considering the width $N$ of the regions of interest, each image frame generates a scan of the filament pseudo-surface, consisting of $N$ individual slices. Figure 10 illustrates the resulting pseudo-surfaces for the normal (top) and abnormal (bottom) sections of the filament.

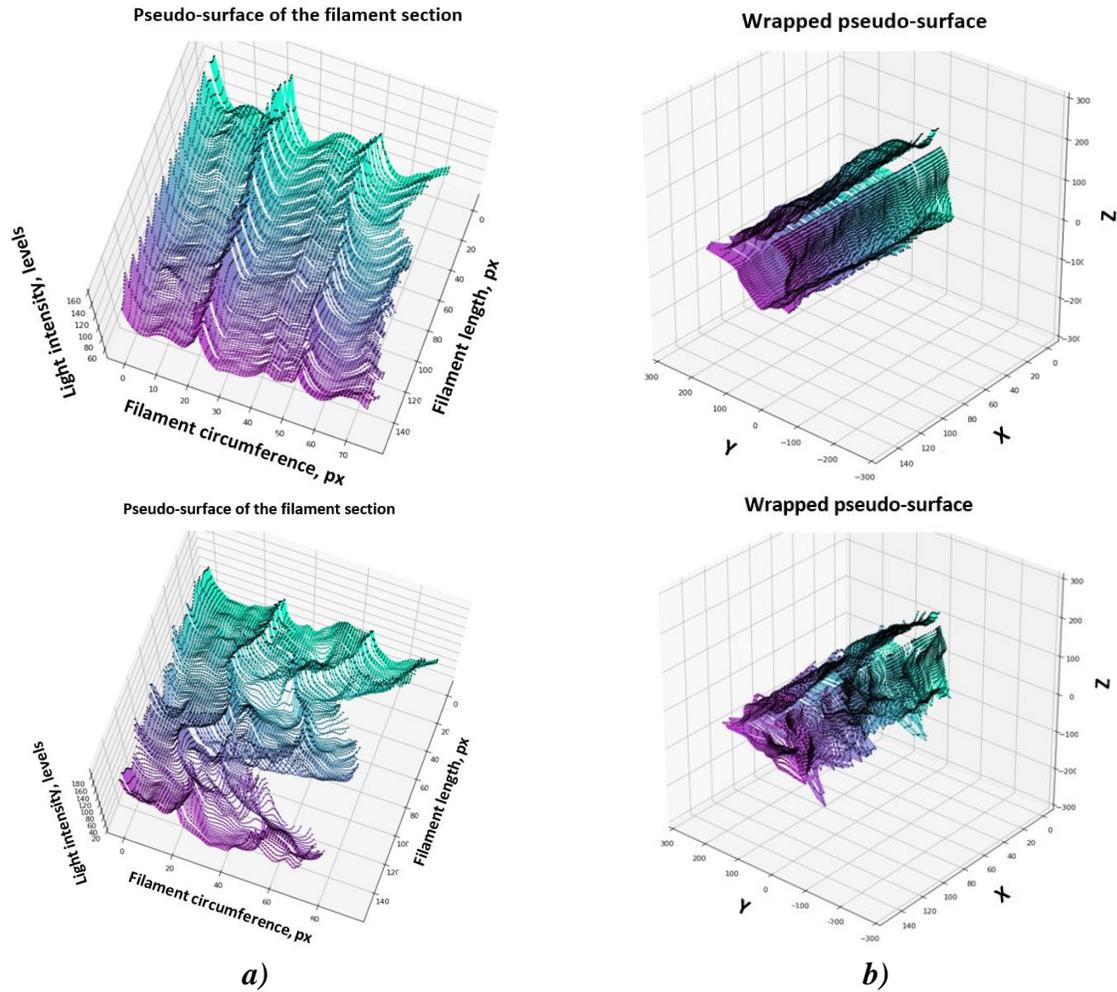

**Figure 10. Pseudo-surface of the filament section obtained in one image frame:** a) light intensity distribution over the entire surface of the filament section; b) wrapped pseudo-surface for the normal (top) and abnormal (bottom) sections of filament.

The intensity of the reflected light cannot reliably convey the real structure of the entire filament exterior, but analysis of its distribution reveals statistical deviations indicating structural anomalies. This analysis, however, requires additional calibrations for different types of plastic and is left for future work.

The developed method makes it possible to track deviations in diameter and ovality of the plastic filament, as well as to estimate the size and location of its structural anomalies.

### 2.3 Calibration

As can be seen from Figure 4, the size of the filament projection depends on its distance to the camera, which is unknown. Having at least one mirror, however, makes it possible to determine the distance between the camera and the extruded filament by the centerline of its mirror projection, which is shown in Figure 11. Therefore, the auxiliary mirrors not only allow performing additional measurements and assessing the ovality of the filament, but also allow for determining the distance to the measured object, which implements an algorithm independent on the horizontal movement of the filament.

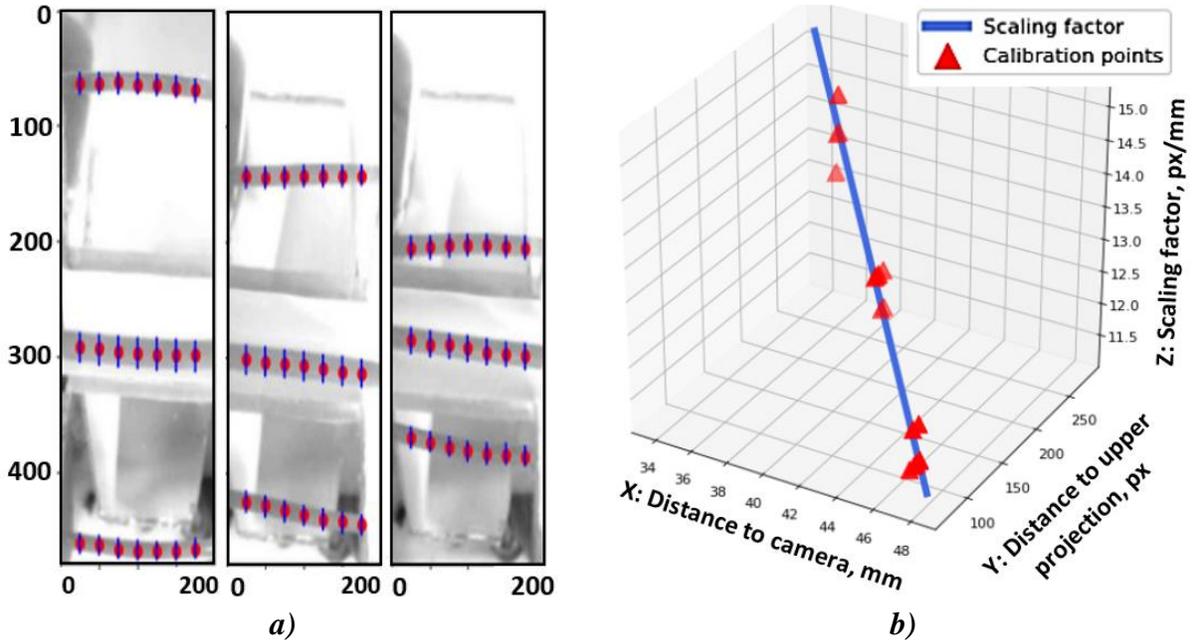

**Figure 11. Determining the distance to the camera by the position of the reflected filament image:** a) Input images with the manually measured diameter and filament location; b) dependence of the scaling factor (in pixels to millimeters) on the distance between the camera and filament (in millimeters), and on the distance between the main image of the filament and its upper mirror projection (in pixels).

The calibration curve matches the number of pixels in the segmented filament image to its actual diameter in millimeters. Due to the recyclebot design, the movement of the filament is limited to the plane perpendicular to the camera image. This simplifies calibration by reducing the number of cases where the filament moves vertically, altering the projected picture.

Starting from finding the centerline of the reflected image, the distance from the filament to the camera is calculated, which makes it possible to apply the appropriate scaling factor and measure the real filament diameter based on the number of segmented pixels (3):

$$\frac{x - x_1}{a} = \frac{y - y_1}{b} = \frac{z - z_1}{c} \tag{3}$$

Where $x$ is the distance between centerlines of the upper mirror projection and the main filament image in pixels, $y$ is the distance between the filament and the camera in millimeters, $z$ is the scaling factor for the diameter measurements in pixels per millimeter, $a$, $b$, and $c$ are the coefficients obtained during calibration, respectively.

Despite the fact that the calibration function is inversely proportional to the distance to the camera and is non-linear, for small filament displacements it can be considered as linear. In the used recyclebot installation, due to the mechanical filament guides, the distance to the camera remains practically unchanged. Thus, the function in Figure 11 is for illustrative purposes only.

To verify and refine the scaling factor of the pixel image of the filament into the physical size in millimeters, a series of experiments were carried out with commercial filament of various diameters. Figures 12 and 13 depict measured diameters of filaments with nominal diameters of 2.90 mm and 1.72 mm, respectively.

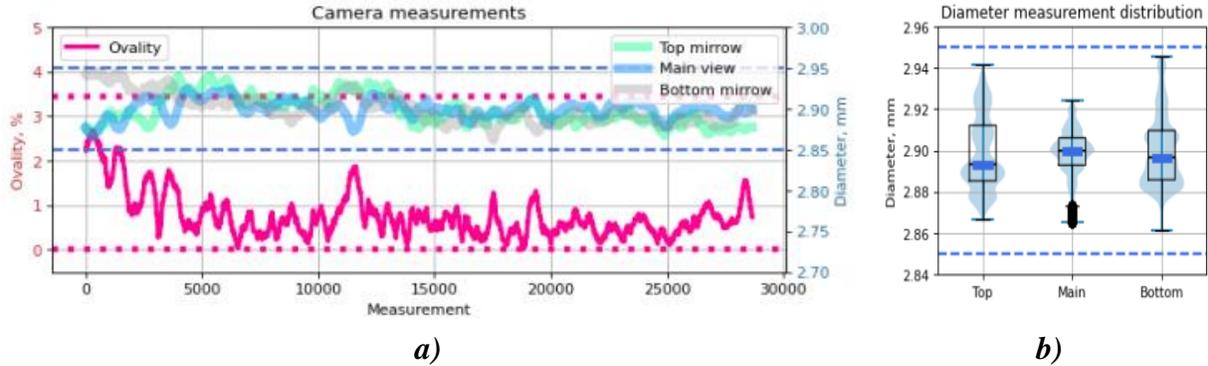

*a)* *b)*

**Figure 12. Measuring the diameter of a commercial 3.0 mm filament:** a) diameter and ovality measurements during the winding process; b) diameter distributions for the main image and upper and lower mirror reflections. Dashed and dotted lines represent the manufacturer's declared diameter and ovality tolerance limits, respectively.

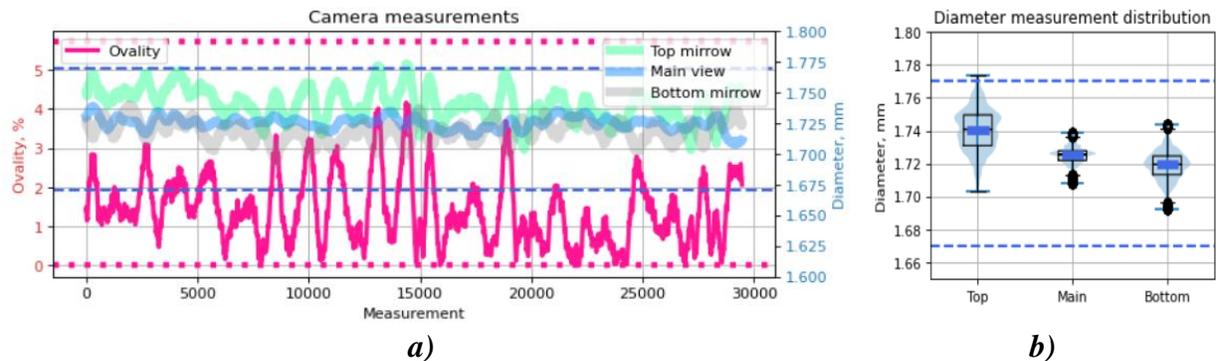

*a)* *b)*

**Figure 13. Measuring the diameter of a commercial 1.75 mm filament:** a) diameter and ovality measurements during the winding process; b) diameter distributions for the main image and upper and lower mirror reflections. Dashed and dotted lines represent the manufacturer's declared diameter and ovality tolerance limits, respectively.

From Figures 12 and 13, it follows that the measured diameter deviation lies within the stated tolerance of ±0.05 mm. The maximum value of ovality, in this case, is *(0.1/$D_N$)· 100 %,* where $D_N$ is the nominal value of the filament diameter.

### 2.4 Experimental Validation on Recycled Materials

During the PLA recycling procedure, several plastic mixtures were prepared based on shredded parts together with virgin PLA pellets. The addition of 25% to 50% virgin PLA granules helps to create a softer texture in printed parts and also increases their tensile strength (3DHubs, 2015) (i.e. counteract the mechanical weakening observed for each recycling cycle (Sanchez, et al. 2017)). In the conducted experiments, the plastic mixture for PLA filament consisted of 70% recycled material and 30% virgin pellets (red, blue, and gray filaments) and 100% recycled material (blue filament).

To create a filament from ABS plastic, only printing products were used, shredded into small granules without using any additional components. Before recycling, each plastic particle mixture was kept separately in a vacuum chamber for 12 hours at a temperature of 70°C to remove adsorbed gases.

Miron et al. (2017) recommended using 175-195°C temperature range for PLA plastic recycling and 165-185°C – for ABS material. Cardona et al. (2016) mentioned the optimal temperature range of 174-180°C for ABS plastic, which depends on the ambient temperature and nozzle diameter. In this work, the nozzle temperature was set at 170°C for PLA plastic and 182°C for ABS plastic with the recycled filament diameter of 2.4-2.8 mm. The ambient temperature ranged from 20°C to 25°C.

Three 400 grams spools of blue, red, and gray PLA plastic, 200 grams of green ABS plastic, and 600 grams of black ABS plastic were produced during testing. From the obtained filament materials ASTM D638-14 type iv tensile bars and ASTM D695 compression test cylinders were fabricated on a LulzBot TAZ 6 (Lulzbot, 2020) 3-D printer with the following printing parameters: print speed of 60 mm/s, line thickness of 0.4 mm, shell thickness of 0.4 mm for PLA and 0.8 mm for ABS, layer thickness of 0.2 mm, and 100% infill. The bars and cylinders were printed using layers with alternating 45- and 135-degree infill lines. The nozzle and bed temperatures were set to 205°C and 60°C, respectively, for PLA plastic, and 235°C and 100°C for ABS plastic.

STL files and printing profiles for the tensile bars and compression cylinders are available in the source file repository (Petsiuk and Pearce, 2020). For each type of plastic, five tensile bars and five compression cylinders were produced.

The tensile and compression tests were performed using an Instron 4206 testing machine along with a 5000 lb Futek load cell (Model LCF455). The extension data was captured by the testing machine based on the crosshead position.

During the tensile tests, the cross-section calculated individually for each tensile bar varied from 24.3 mm$^2$ to 29.8 mm$^2$. The peak load, in turn, was 1.29±0.15 kN (1.45±0.05 kN for samples made from commercial plastic) for PLA specimens and 0.68±0.29 kN (0.92±0.01 kN for samples made from commercial plastic) for ABS specimens. For compression tests, the cross-sectional area was considered unchanged and equal to 176.7 mm$^2$. Compression stress was used as a metric, which is the pressure required to compress the sample by 1 mm. The average values of strength and stress for each set of samples and their comparative analysis are presented below.

For each type and color of recycled plastic, two diameter measurement tests were performed, the results of which were compared to manual measurements obtained with a Mitutoyo 500-196-30 digital caliper with a resolution of 0.01 mm, as well as with the measurement results using the one-dimensional light sensor (Woern et al., 2018a). Since the plastic feed rate specified by G-Code commands during printing is based on the diameter of the filament, a comparison of the calculated and actual weights of manufactured parts could also be considered as an indirect indicator of the accuracy of the filament diameter measurements (Tanikella, et al., 2017).

## 3. Results

Table 2 shows the sequence for obtaining spools of recycled plastic filament starting with the selection of used printed parts.

**Table 2. Recycled PLA and ABS filament.**

| | Recycled material | Pelletized mixture | Recycled filament |
|---|---|---|---|
| | **PLA** | | |
| 1 | 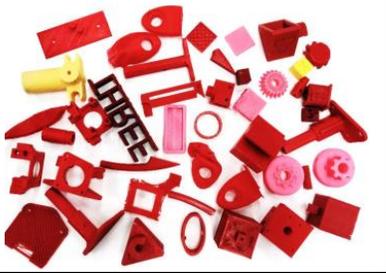 | 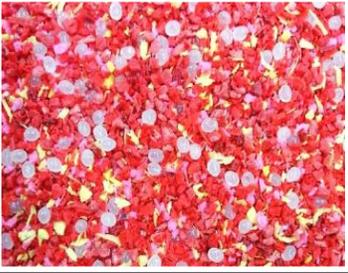 | 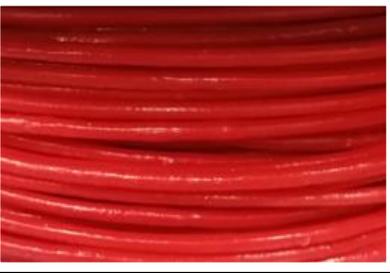 |
| 2 | 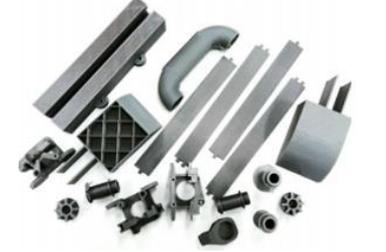 | 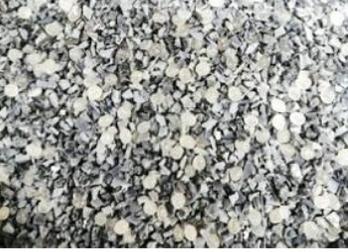 | 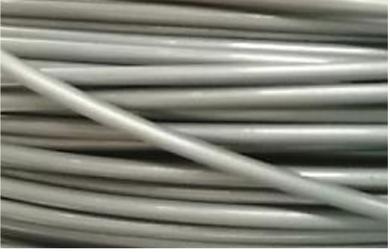 |
| 3 | 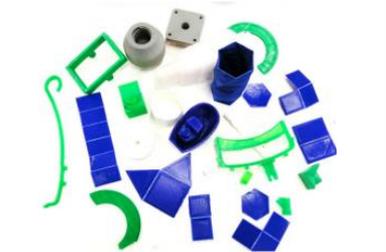 | 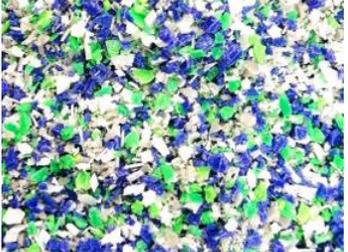 | 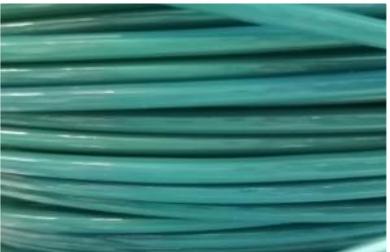 |
| | **ABS** | | |
| 4 | 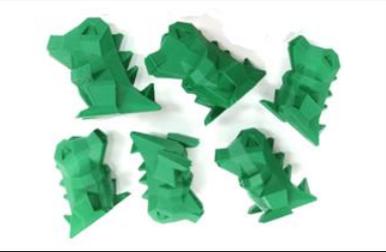 | 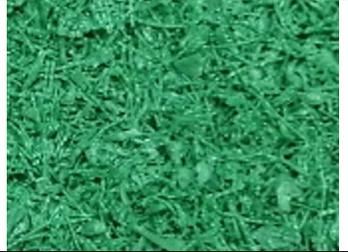 | 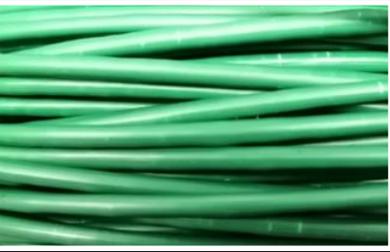 |
| 5 | 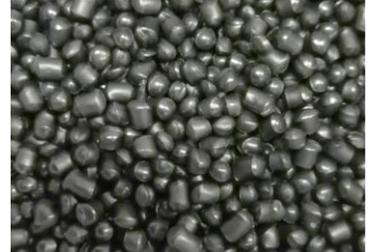 | 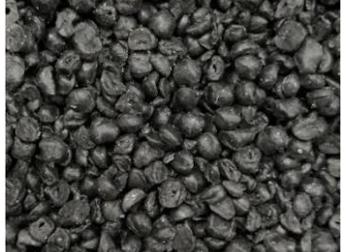 | 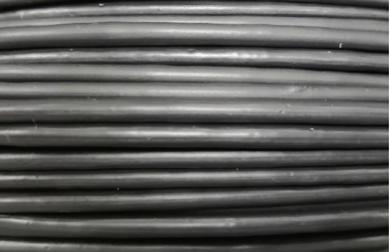 |

Figures 14 and 15 show the distribution of measured filament diameters and ovality for different types and colors of the recycled plastic mixtures.

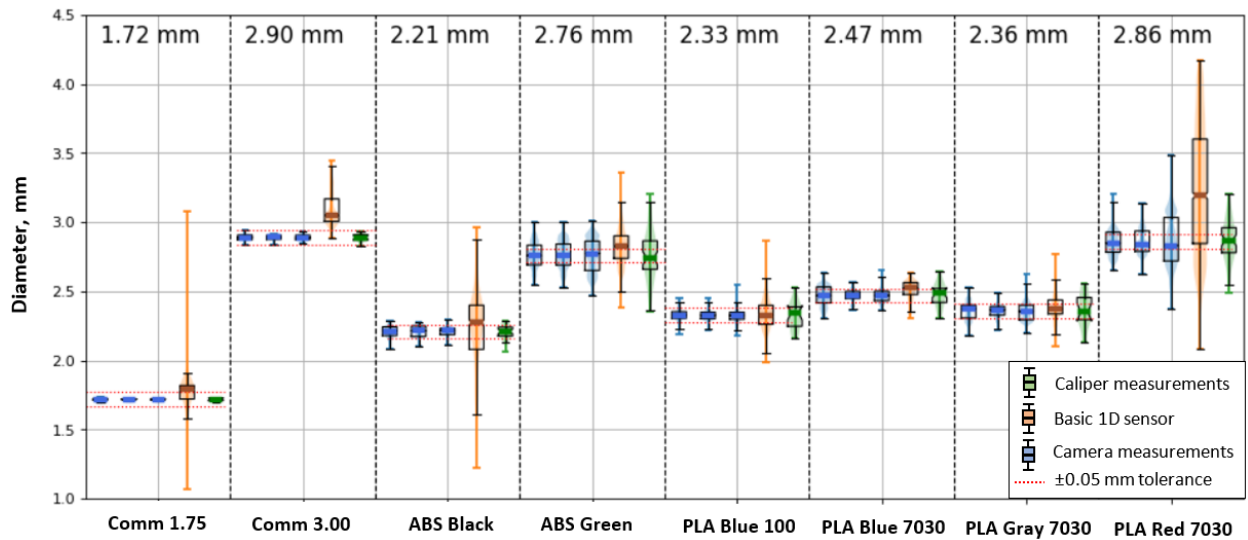

**Figure 14. Distribution of diameters for each produced filament obtained using different measurement methods.**

As can be seen in Figure 14 the camera measurements provided by the sensor system developed in this study are far superior to the basic 1-D sensor for all filaments investigated, and comparable to those obtained with the digital caliper measurements. The sensor system here not only measures the diameter, but also calculates ovality - key to determining print quality and, in the worst cases, printability itself.

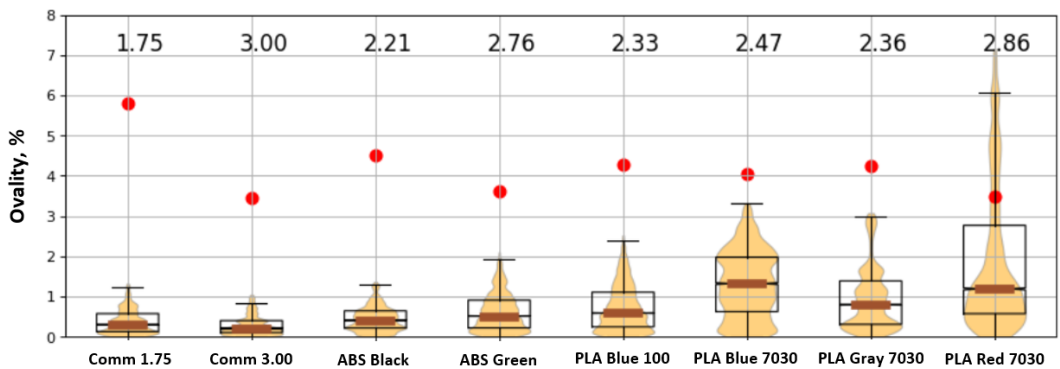

**Figure 15. Distribution of ovality values obtained by optical diameter measurements with the camera. Red dots represent the maximum allowable ovality for each diameter.**

From the produced plastic spools, tensile bars and compression cylinders were manufactured for the corresponding tests. The quality of the red PLA plastic was not good enough to print testing samples (as could be predicted by its large deviations in ovality observed in Figure 15). Figure 16 shows the mass distribution of the fabricated tensile bars and compression cylinders with indicated average tensile strength and compressive stress values.

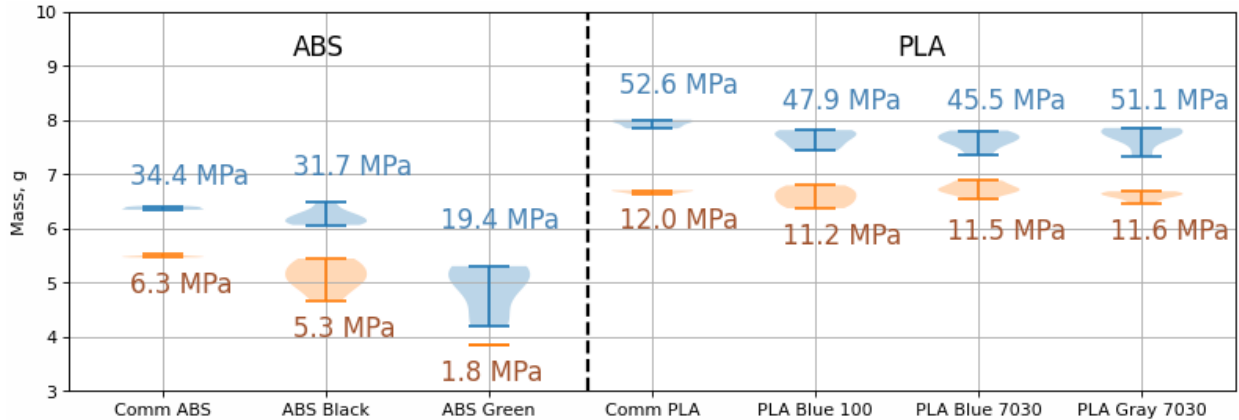

**Figure 16. Mass distribution of the produced tensile bars (blue) and compression cylinders (orange) with indicated average tensile strength (blue) and compressive stress (orange) values.**

It should be noted, that the green ABS plastic had one compression cylinder so is not as reliable as all of the other measurements. As can be seen from Figure 16, the strength of recycled plastic determined from 5 measurements per sample is approximately 85-97% of the strength of virgin plastic for PLA, which had some virgin PLA added to the mixture to maintain strength. For the ABS, which was processed directly the strength was 56-92% of virgin materials. This is consistent with previous experiments (Lanzotti et al., 2019). The strength values of reference specimens made from commercial plastic spools also match other similar tests (Laureto and Pearce, 2018).

## 4. Discussion and future work

The results here show that the diameter sensor is effective for two of the most popular DRAM plastics. The 3-D printing technical community has embraced open source methods to recycle 3-D printing waste (Hunt et al., 2015) not only for PLA (Sanchez et al., 2015) (including virgin, recycled (Anderson, 2017) and biodegrable PLA (Pakkanen et al., 2017)) and ABS whether with for humanitarian aid (Mohammed et al., 2018a) or sustainable manufacturing (Mohammed et al., 2018b)). It is clear other common post-consumer thermoplastics need to be studied in future work. Already plastic waste has been upcycled with 3-D printing including polyethylene terephthalate (PET) (Zander et al., 2018) and recycled and virgin PET blends (Lee et al., 2013), high density polyethylene (HDPE) (Chong et al., 2017), polypropylene (PP) and polystyrene (PS) (Pepi et al., 2018), thermoplastic polyurethane (TPU) (Woern and Pearce, 2017), linear low density polyethylene (LLDPE) and low density polyethylene (LDPE) (Hart et al., 2018), polycarbonate (PC) (Reich et al., 2019) as well as blends (Zander et al., 2019).

In addition, as each melt and extrude cycle of a recyclebot impairs the mechanical properties of PLA (Sanchez et al., 2017), HDPE (Oblak et al., 2015), and (PET) (Lee et al., 2013), this limits the recycling cycles to approximately five (Sanchez et al., 2017). Thus, to reach the goals of the circular economy reinforcement or blending with virgin materials becomes necessary. Thus blends, as this study has demonstrated for PLA, should also be investigated in future work. Even indications from multiple cycles of ABS (Mohammed et al., 2019) and accelerated aging (Boldizar and Möller, 2003) indicate such techniques will be necessary for the other thermoplastics as well.

In addition, polymer composites using carbon-reinforced plastic (Tian et al., 2017), fiber-filled composites (Parandoush and Lin, 2017) particularly for large areas (Heller et al., 2019), and various types of waste wood (Pringle et al., 2017) and wood flour composites (Zander, 2019) have been used in recyclebot systems. These filaments are not uniform and their use may be increased with the information provided by the pseudo-surface of the filament section made possible with the new system described in this study. For example, future work could investigate the improved print quality and strength made possible by using this sensor on the input of FFF 3-D printer to adjust the print parameters in real time for deviations in the filament following Greeff and Schilling (2017). With the system described here when measuring the filament's diameter in real time as it is fed into the hot end of a 3-D printer, the expected diameter and can be divided by the measured value to get a percentage. That percentage can then be used as the extruder multiplier, which slows or speeds up the stepper motor attached to the extruder as needed. Such automation around bad filament has been demonstrated in the past for simple sensors to greatly improve print quality (Coetzee, 2016). Substantial future work is needed in this area. In addition, it should also be noted that the developed system is highly sensitive even to small displacements of the camera and mirrors, which requires periodic recalibration. Future work can investigate a method to automate this task.

The open source filament diameter sensor is more sophisticated and accurate that other systems yet costs less and is easy to integrate into an Arduino controlled recyclebot system as well as for FFF 3-D printing. These opens up the business opportunity for small and medium sized enterprises to commercial filament and compete with major manufactures. There is already a market for recycled filament. For example, a few companies sell recycled PET filament including Refil and B-PET.

## 5. Conclusions

The developed open source filament sensing method allows users to obtain significantly more information in comparison with basic one-dimensional light sensors and using the received data not only for more accurate diameter measurements, but also for a detailed analysis of the recycled filament surface. Experiments with commercially produced plastic spools have shown measurement uncertainties within the standard tolerance of ±0.05 mm, indicating sufficient accuracy and reliability. This technique can be used with any types and colors of plastic, however, if the filament color is close to the background tone, it may be necessary to limit the regions of interest. In the future, the optical module can be integrated into both the recyclebot and 3-D printer control systems to regulate their motors in real-time based on the measured filament parameters.

## 6. Acknowledgements

This work was supported by the Witte Endowment. The authors would like to acknowledge helpful discussions with Adam Pringle and Nagendra Tanikella.